\documentclass[conference]{IEEEtran}
\IEEEoverridecommandlockouts
\usepackage{cite}
\usepackage{lipsum}
\usepackage{amsmath,amssymb,amsfonts}
\usepackage{algorithmic}
\usepackage{algorithm}
\usepackage{graphicx}
\usepackage{textcomp}
\usepackage{tabularx}
\usepackage{booktabs}
\usepackage{graphicx}
\usepackage{caption}
\usepackage{tabularx}
\usepackage{booktabs}
\usepackage[table,xcdraw]{xcolor}
\usepackage{pifont}
\usepackage{amssymb}
\usepackage{pifont}
\newcommand{\mycomment}[1]{}
\def\BibTeX{{\rm B\kern-.05em{\sc i\kern-.025em b}\kern-.08em
    T\kern-.1667em\lower.7ex\hbox{E}\kern-.125emX}}

\definecolor{LightBlue}{rgb}{0.85, 0.92, 1.0}
\definecolor{HeaderBlue}{rgb}{0.7, 0.85, 1.0}

\definecolor{LightBlue}{rgb}{0.85, 0.92, 1.0}
\definecolor{HeaderBlue}{rgb}{0.7, 0.85, 1.0}
\definecolor{MyGreen}{rgb}{0, 0.5, 0}
\definecolor{MyRed}{rgb}{0.6, 0, 0}
\definecolor{MyGray}{rgb}{0, 0, 0}
\rowcolors{2}{LightBlue}{white}
\usepackage{orcidlink}

\newcommand{\orcidauthorA}{0009-0002-4216-8912} 
\newcommand{\orcidauthorB}{0009-0009-4851-2666} 
\newcommand{\orcidauthorC}{0009-0008-7865-0544} 
\newcommand{\orcidauthorD}{0000-0002-8621-6210} 

\begin{document}

\title{A System for Comprehensive Assessment of RAG Frameworks\\


\thanks{}
}

\author{
\IEEEauthorblockN{
    Mattia Rengo\textsuperscript{\large{{\orcidlink{\orcidauthorA}}}}, 
    Senad Beadini\textsuperscript{\large{{\orcidlink{\orcidauthorB}}}}, 
    Domenico Alfano \textsuperscript{\large{{\orcidlink{\orcidauthorC}}}},
    Roberto Abbruzzese \textsuperscript{\large{{\orcidlink{\orcidauthorD}}}}
}
\IEEEauthorblockA{
\textit{R\&D Department, Eustema S.p.A.}\\
Napoli, Italy \\
\{m.rengo, s.beadini, d.alfano, r.abbruzzese\}@eustema.it
}
\\
\IEEEauthorblockA{
\textbf{Code}: \text{\href{https://github.com/Eustema-S-p-A/SCARF}{https://github.com/Eustema-S-p-A/SCARF}}
}

}

\maketitle

\begin{abstract}
Retrieval Augmented Generation (RAG) has emerged as a standard paradigm for enhancing the factual accuracy and contextual relevance of Large Language Models (LLMs) by integrating retrieval mechanisms. However, existing evaluation frameworks fail to provide a holistic black-box approach to assessing RAG systems, especially in real-world deployment scenarios. To address this gap, we introduce SCARF (System for Comprehensive Assessment of RAG Frameworks), a modular and flexible evaluation framework designed to benchmark deployed RAG applications systematically. SCARF provides an end-to-end, black-box evaluation methodology, enabling a limited-effort comparison across diverse RAG frameworks. Our framework supports multiple deployment configurations and facilitates automated testing across vector databases and LLM serving strategies, producing a detailed performance report. Moreover, SCARF integrates practical considerations such as response coherence, providing a scalable and adaptable solution for researchers and industry professionals evaluating RAG applications. Using the REST APIs interface, we demonstrate how SCARF can be applied to real-world scenarios, showcasing its flexibility in assessing different RAG frameworks and configurations.
SCARF is available at \href{https://github.com/Eustema-S-p-A/SCARF}{GitHub repository}.
\end{abstract}


\section{Introduction}

Retrieval-Augmented Generation (RAG) represents a remarkable advancement in Natural Language Processing (NLP), significantly enhancing the performance of generative Language Models (LMs). By combining the capabilities of LMs with external knowledge bases, RAG allows responses that are not only more accurate but also contextually relevant. Merging information retrieval with language generation, RAG systems address a major limitation of standalone LMs: their tendency to generate responses that, while coherent, may lack factual accuracy or grounding.
Since traditional Large Language Models (LLMs) rely solely on pre-trained data, they may generate factually incorrect information and unreliable outputs. RAG systems address this limitation by retrieving relevant information in real-time, making them well-suited for tasks that require up-to-date, accurate, and context-aware responses, such as answering questions, generating content, and supporting various real-world applications.
RAG approaches \cite{gao2023retrieval} have evolved rapidly, resulting in a wide variety of system variants \cite{asai2023self, trivedi2022interleaving, borgeaud2022improving, deng2023regavae} and benchmarking frameworks \cite{yu2024evaluation}. Many existing evaluation methods \cite{zhu2024rageval, langchainbenchamrk} focus on assessing specific components, such as the relevance of retrieved documents or the quality of generated responses. However, these methods often lack a comprehensive perspective, failing to provide a holistic, end-to-end evaluation that considers not only the interplay between the retrieval and generation components but also the flexibility to experiment with and optimize the underlying technical tools used in these processes. Another critical yet often overlooked aspect is the variability introduced by different deployment frameworks used for LLMs. Tools such as vLLM \cite{kwon2023efficient}, OpenLLM \cite{openllm}, and Ollama \cite{ollama} implement diverse optimization strategies, including techniques like quantization, batching, and caching. These strategies can have a significant impact on key performance metrics, such as latency, system efficiency, and overall responsiveness. These variations are particularly important in real-world scenarios, where deployment choices can directly affect the user experience and system scalability. Despite their importance, existing evaluation methods rarely integrate the ability to manage and measure the influence of these deployment frameworks systematically.
To address these limitations, we propose SCARF, a low-level evaluation framework designed to assess RAG systems comprehensively while maintaining extreme modularity. This framework, in addition to offering a wide range of capabilities, also enables an easy replacement of individual system components, such as testing various vector databases for the retrieval phase or employing different LLM deployment frameworks in the generation phase. By providing this flexibility, our framework supports targeted experiments to understand the impact of specific system changes, including deployment strategies and their optimizations.
Another strength of our approach is its ability to generate a detailed final report that provides test results of the RAG pipeline. This report allows for precise measurement of system performance in specific scenarios, such as addressing particular types of questions or solving specialized tasks.
Moreover, our evaluation framework extends beyond traditional metrics by leveraging state-of-the-art LLM-based evaluation methods \cite{zheng2023judging}, including RAGAS \cite{es2023ragas}.
To summarize, our framework offers the following key advantages:

\begin{enumerate}
    \item \textbf{Easy to integrate and use}: Researchers and practitioners can set up and evaluate RAG systems with minimal configuration.
    \item \textbf{Highly modular and customization}: Our framework supports modular component replacement, enabling systematic experimentation.
    \item \textbf{Comprehensive and insightful}: By producing a final report, the framework enables a deep understanding of system performance. The report provides valuable insights that can guide to hyper-params tuning and improvements for specific use cases or scenarios.
\end{enumerate}

\section{Background and Related Work}

\begin{table*}[ht]
    \caption{Comparison of Frameworks. This comparison highlights the strengths and limitations of each framework, providing insight into their capabilities and identifying areas where improvements or extensions may be needed. The table presents an overview of the different frameworks and their support for key features. A {\color{MyRed}{\ding{55}}} symbol indicates that the feature is not implemented in the respective framework. A {\color{MyGreen}{\ding{51}}} signifies full support, meaning the feature is fully integrated and functional without limitations. The {\color{MyGray}\(\circleddash\)} symbol represents partial support, meaning the feature is present but lacks completeness. }
    \label{tab:framework_comparison}
    \centering
    \renewcommand{\arraystretch}{1.2} 
    \setlength{\tabcolsep}{5pt}       
    \small                            
    \begin{tabular}{>{\centering\arraybackslash}m{3cm} 
                    >{\centering\arraybackslash}m{1.6cm} 
                    >{\centering\arraybackslash}m{1.8cm} 
                    >{\centering\arraybackslash}m{1.6cm} 
                    >{\centering\arraybackslash}m{1.6cm} 
                    >{\centering\arraybackslash}m{1.6cm} 
                    >{\centering\arraybackslash}m{1.6cm}
                    >{\centering\arraybackslash}m{1.6cm}} 
        \toprule
        \rowcolor{HeaderBlue} 
        \textbf{Framework} & \textbf{Retrieval Metrics} & \textbf{Generation Metrics} & \textbf{Synthetic Data Gen} & \textbf{Multi-RAG Testing} & \textbf{External RAG Support} & \textbf{Config \& Auto Testing} & \textbf{API Integration} \\ 
        \midrule
        
        Langchain Bench \cite{langchainbenchamrk} & {\color{MyGray}\(\circleddash\)} & {\color{MyGreen}\ding{51}} & {\color{MyRed}\ding{55}} & {\color{MyGreen}\ding{51}} & {\color{MyRed}\ding{55}} & {\color{MyRed}\ding{55}} & {\color{MyRed}\ding{55}} \\ 
        RAG Evaluator \cite{RAGEvaluator}   & {\color{MyGray}\(\circleddash\)} & {\color{MyGreen}\ding{51}} & {\color{MyRed}\ding{55}} & {\color{MyRed}\ding{55}} & {\color{MyRed}\ding{55}} & {\color{MyRed}\ding{55}} & {\color{MyRed}\ding{55}} \\ 
        Giskard (RAGET) \cite{giskard} & {\color{MyGray}\(\circleddash\)} & {\color{MyGreen}\ding{51}} & {\color{MyGreen}\ding{51}} & {\color{MyRed}\ding{55}} & {\color{MyRed}\ding{55}} & {\color{MyRed}\ding{55}} & {\color{MyRed}\ding{55}} \\ 
        Rageval  \cite{zhu2024rageval}        & {\color{MyGreen}\ding{51}} & {\color{MyGreen}\ding{51}} & {\color{MyRed}\ding{55}} & {\color{MyRed}\ding{55}} & {\color{MyRed}\ding{55}} & {\color{MyRed}\ding{55}} & {\color{MyRed}\ding{55}} \\ 
        Promptfoo \cite{promptfoo}       & {\color{MyRed}\ding{55}} & {\color{MyGreen}\ding{51}} & {\color{MyRed}\ding{55}} & {\color{MyRed}\ding{55}} & {\color{MyRed}\ding{55}} & {\color{MyRed}\ding{55}} & {\color{MyRed}\ding{55}} \\ 
        ARES \cite{saad2023ares}  & {\color{MyGreen}\ding{51}} & {\color{MyGreen}\ding{51}} & {\color{MyGreen}\ding{51}} & {\color{MyGray}\(\circleddash\)} & {\color{MyRed}\ding{55}} & {\color{MyGreen}\ding{51}} & {\color{MyRed}\ding{55}} \\ %
        
        \textbf{SCARF} (ours)     & {\color{MyGreen}\ding{51}} & {\color{MyGreen}\ding{51}} & {\color{MyRed}\ding{55}} & {\color{MyGreen}\ding{51}} & {\color{MyGreen}\ding{51}} & {\color{MyGreen}\ding{51}} & {\color{MyGreen}\ding{51}} \\ 

        \bottomrule
    \end{tabular}
\end{table*}

The evaluation of LLMs and RAG systems necessitates specialized frameworks capable of assessing both retrieval performance—such as relevance and recall—and generation quality, including coherence and factual accuracy. In this section, we critically examine representative solutions, emphasizing their approaches to data handling, metric evaluation, and deployment scenarios. \\
Evaluating RAG systems is a complex task due to their dual nature of retrieving relevant information and generating coherent responses \cite{yu2024evaluation}. Traditional evaluation methods for language models, such as perplexity or human annotations, are often inadequate for capturing the nuanced interplay between these two components. For example, specialized metrics are needed to assess how effectively the retrieved context supports the generation process and ensures the accuracy and relevance of the output. Recent research has increasingly leveraged LLM-as-a-judge \cite{zheng2023judging} techniques to develop advanced evaluation metrics for retrieval-augmented generation pipelines. \cite{fu2023gptscore} proposed GPT-Score that leverages the generative capabilities of pre-trained language models—such as GPT-3 \cite{brown2020language} or GPT-4 \cite{achiam2023gpt}—to assess the quality of a generated text. RAGAS \cite{es2023ragas} further refines these approaches by defining three specialized metrics specifically designed for RAG applications: faithfulness, answer relevance, and context relevance.

\textbf{Faithfulness} \cite{es2023ragas} measures the consistency of the generated response with the retrieved context. It determines whether the claims made in the response are substantiated by the retrieved passages. A response that aligns closely with the context receives a high faithfulness score, while deviations or hallucinations result in lower scores. Within RAGAS, faithfulness is assessed using a structured scoring system, allowing evaluators—both human and automated—to rate the degree of alignment between the generated content and the supporting context. This metric is critical to ensuring that RAG systems produce factual and trustworthy outputs.

\textbf{Answer relevancy} \cite{es2023ragas} evaluates how well the generated response addresses the user’s query. This metric is essential for ensuring that the response is not only accurate but also directly relevant to the user’s needs. In the RAGAS framework, answer relevance is assessed by analyzing the relationship between the query’s intent and the content of the response. High scores in answer relevance reflect responses that provide meaningful and precise information, making this metric central to determining the usefulness of RAG outputs.

\textbf{Context relevancy} \cite{es2023ragas} focuses on the quality and specificity of the information retrieved by the system. It examines whether the retrieved passages are relevant to the input query and sufficiently focused to support accurate and coherent response generation. High context relevance scores indicate that the retrieved information is appropriate and effectively contributes to the final response. This metric is pivotal for evaluating the performance of the retrieval component within RAG systems, as it directly influences the overall quality of the output.

Other common metrics in NLP, such as ROUGE \cite{lin2004rouge} and BLEU \cite{papineni2002bleu}, can be applied to evaluate certain properties of RAG-generated outputs. These metrics specifically measure the overlap between generated outputs and reference answers, making them suitable for assessing lexical similarity. However, their reliance on overlap-based evaluation limits their ability to capture other aspects of RAG performance, which are better addressed by the specialized metrics provided by RAGAS.

\subsection{Key Feature of RAG Evaluators}

In this section, we systematically review and analyze the essential characteristics that any framework designed to evaluate framework RAG systems should possess. Our objective is to dissect the key features of these tools and elucidate how they overcome the challenges of assessing RAG's retrieval and generation components. 
The main features are presented in Table~\ref{tab:framework_comparison}. Below, we provide a brief explanation of each column:

\begin{itemize}
\item \textbf{Retrieval Metrics:} indicates whether the framework supports the evaluation of the retrieval phase by measuring metrics such as recall, precision, or relevance of the retrieved documents. Tools that can directly score this stage of an RAG pipeline or generate retrieval-centric metrics are noted here. 
\item \textbf{Generation Metrics:} reflects the framework's capability to assess the generated text's quality. This includes standard metrics (e.g. BLEU \cite{papineni2002bleu}, ROUGE\cite{lin2004rouge}) as well as specialized LLM-based metrics.
\item \textbf{Synthetic Data Gen:} denotes whether the tool is equipped to automatically generate new test data or augment existing datasets—typically by creating new question-answer pairs from a given knowledge base. This feature helps extend evaluations to new domains without requiring extensive manual annotation. 
\item \textbf{Multi-RAG Testing:} specifies if the tool can simultaneously compare or evaluate multiple RAG solutions. This multi-system testing allows for side-by-side benchmarking under consistent conditions, thereby facilitating comprehensive performance comparisons. \item \textbf{External RAG Support:} determines whether the framework is capable of interfacing with fully deployed, third-party RAG endpoints. We call this also “black-box” testing capability. This feature is essential when direct access to internal components is not feasible. 
\item \textbf{Config \& Auto Testing:} indicates if the framework supports a configuration-based or script-based approach that enables automatic execution of tests—including tasks such as data uploads and query submission—without requiring extensive manual intervention. 
\item \textbf{API Integration:} this column highlights whether the framework is designed to integrate via standard APIs, thus enhancing interoperability and efficiency in real-world deployments. In industrial settings, systems are frequently exposed through standardized APIs (e.g., REST), thus this feature may be essential for the usability of a framework.
\end{itemize}

\subsection{Related Frameworks}

In this section, we examine several promising tools for RAG evaluation, assessing each one based on the key features described in the preceding discussion. By systematically analyzing how each tool addresses these criteria, we highlight their respective strengths, limitations, and potential gaps—thereby providing a clear perspective on the current state of RAG evaluation practices.\\
\textbf{RAG evaluator} \cite{RAGEvaluator} is a Python library that supports a broad set of text-generation metrics like BLEU \cite{papineni2002bleu}, ROUGE \cite{lin2004rouge}, Bert-Score \cite{zhang2019bertscore}, METEOR \cite{banerjee2005meteor} and can detect certain bias or hate-speech aspects. This makes it straightforward to evaluate generation quality on a precollected dataset. However, it does not inherently integrate a procedure for evaluating external RAG systems; users typically provide the generated text and references offline rather than hooking into a live RAG framework deployment.\\
\textbf{Giskard} \cite{giskard} offers a comprehensive testing and scanning approach for AI systems, including LLMs and RAG. Its RAG Evaluation Toolkit (RAGET) can auto-generate synthetic question-and-answer pairs from a knowledge base, perform correctness checks, and detect harmful or biased content. However, it primarily assumes local or in-house RAG integration, focusing on diagnosing and improving the pipeline rather than systematically comparing different third-party RAG frameworks.
\textbf{Rageval} \cite{zhu2024rageval} is a fine-grained evaluator that splits RAG into multiple subtasks (query rewriting, document ranking, evidence verification, etc.), each with dedicated metrics. It can distinguish between “answer correctness” (comparing to ground truth) and “groundedness” (checking alignment with retrieved passages). Like RAG Evaluator, it is primarily used offline and may require additional scripts to interact with a deployed RAG endpoint.
\textbf{Promptfoo} \cite{promptfoo} is a developer-focused CLI that emphasizes rapid iteration over prompts, comparing outputs from different LLMs or prompt variants. It includes a “red teaming” mode to detect potential vulnerabilities (e.g. prompt injections), making it valuable for improving prompt design. Promptfoo, however, does not offer built-in retrieval metrics and is not designed for direct integration with large-scale, black-box RAG frameworks.
\textbf{Langchain Benchmark} \cite{langchainbenchamrk} is an open-source tool designed to evaluate various tasks related to LLMs. Organized around end-to-end use cases, it heavily utilizes LangSmith for its benchmarking processes. Nevertheless, it primarily focuses on local development environments and does not facilitate RAG external black-box testing. Consequently, users must set up and manage evaluations within their own infrastructure, and the benchmarking process is not fully automated.

Table~\ref{tab:framework_comparison} provides a comprehensive comparison of the existing solutions across the key characteristics discussed in the previous section. While each tool has distinct strengths—focusing on specific aspects of the RAG pipeline—there exists a significant gap in the landscape for a more unified, low-level “black-box” evaluator. Such an evaluator should facilitate: 
\begin{itemize} 
\item Seamless interaction with fully deployed, third-party RAG services via APIs, encompassing essential capabilities such as file uploads, query handling, and inference evaluation. 
\item Effortless flexibility in replacing essential components—including vector databases and LLM serving engines (e.g. Ollama, vLLM)—or even entire RAG frameworks, all with minimal configuration overhead and no disruption to the overall evaluation process. 
\item The ability to provide systematic, granular metrics that thoroughly assess the entire RAG pipeline, encompassing critical aspects such as also response latency and resource consumption. 
\end{itemize}

%

\section{Description of the Framework}

In this section, we discuss SCARF, our evaluation framework designed to address existing limitations in RAG evaluators. At its core, SCARF offers a modular Python-based suite that treats RAG platforms as interchangeable "plugins", allowing comprehensive end-to-end evaluations in realistic deployment scenarios. We begin by describing the architecture of SCARF, highlighting its main components and how they interoperate to support flexible testing. Next, we provide a step-by-step guide on using SCARF, illustrating how it can be leveraged into existing pipelines to evaluate RAG systems. This design enables researchers and practitioners to fill the current gaps in the evaluation of RAG by facilitating more scalable and adaptable testing on diverse platforms and configurations.
\subsection{Framework Architecture}
\begin{figure}[htbp]
\centering
\includegraphics[width=.98\columnwidth]{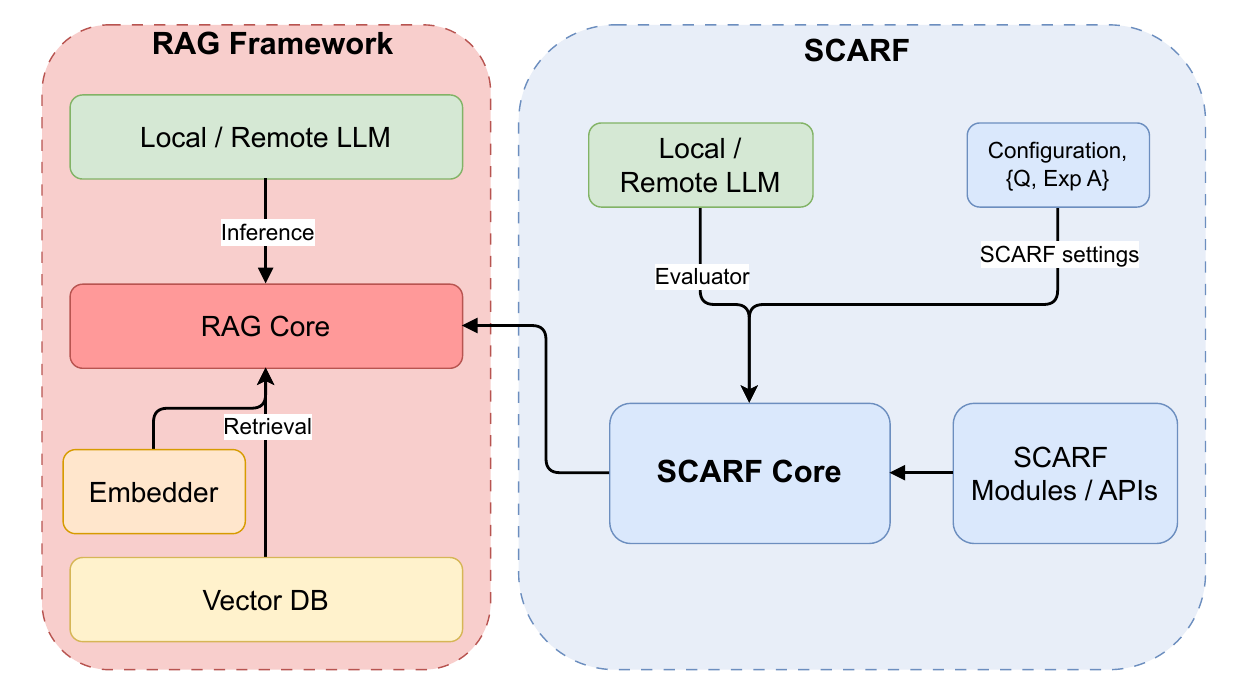}
    \caption{High-level SCARF architecture showing modular integration points for RAG frameworks, vector databases, and LLM engines.}
    \label{fig:scarf_arch}
\end{figure}

SCARF is designed following a \emph{plug-and-play} principle, enabling systematic evaluation of different deployed platforms \emph{without modifying their core implementation}. Figure~\ref{fig:scarf_arch} clearly illustrates the architecture, detailing the distinct functional blocks organized within the project's repository. The repository is structured into four primary sections. The SCARF Core includes core testing scripts and configuration files that specify necessary test datasets and queries required for conducting evaluations. In addition, SCARF Modules and APIs are provided, containing dedicated API adapter modules specifically tailored for integration with various RAG platforms. Another part of the repository consists of Docker Compose files and configuration resources necessary for deploying supported RAG frameworks and for integrating both local and remote LLM engines or vector databases. Although these resources are provided, SCARF does not automatically manage the deployment of these services. Lastly, a dedicated Configuration Settings area provides comprehensive settings and configurations, enabling users to easily manage different evaluation scenarios and customize the evaluation processes according to their specific testing requirements.

\mycomment{
Inside the project's repository, users can find four main folders, each one representing the folder of one of the main blocks depicted in Figure~\ref{fig:scarf_arch} : 
\begin{itemize}
    \item \texttt{frameworks-test/}: Contains core testing scripts, such as \texttt{test\_rag\_frameworks.py}, a \texttt{config.json} for specifying test datasets and queries, as well as various RAG frameworks' API adapter modules (e.g. \texttt{cheshirecat\_api.py}).
    \item \texttt{frameworks-rag/}: Houses Docker Compose files and similar configurations for each supported RAG framework (e.g. \texttt{cheshirecat}, \texttt{anythingllm}). While SCARF currently does not automatically manage these services, users can find and store relevant setup resources here.
    \item \texttt{llm-local-providers/}: Provides configuration and resources for local LLM engines (e.g. \texttt{ollama/compose.yml}).
    \item \texttt{vectorDB-local-providers/}: Contains local vector DB solutions (e.g. Qdrant \cite{qdrant}) and Docker Compose files to set them up.
\end{itemize}
}

Thus SCARF, highlighting its ability to:

\begin{itemize}
    \item \textbf{Interact with multiple frameworks as black boxes:} users can test RAG frameworks (e.g. AnythingLLM \cite{anything-llm} , CheshireCat \cite{cheshire-cat-ai}) \emph{without having to replicate or fully understand their internal workings}, simply by wrapping them with a module that exposes consistent methods for uploading data and querying if such adapter is not already provided by SCARF.
    
    \item \textbf{Leverage different vector databases:} SCARF supports quick reconfiguration of the underlying vector database when the remote RAG  permits such changes. Thanks to the  \texttt{vectorDB-local-providers}, users can seamlessly switch to any local vector database (e.g. Qdrant \cite{qdrant}, Milvus \cite{milvus}) with minimal updates to a Docker Compose file. This ensures minimal overhead when adapting to different storage solutions and maintains SCARF’s goal of providing a flexible and extensible evaluation environment.
    \item \textbf{Use local or remote LLM providers:} within \texttt{llm-local-providers}, users have access to all necessary components for executing local LLM inference using engines such as Ollama, vLLM, or alternative implementations with Docker Compose. Additionally, they can configure a remote API (e.g., OpenAI, Anthropic, OpenRouter) to interface with their preferred models.
    \item \textbf{Test and compare frameworks:} a single Python entry point can spin up multiple tests, collecting results and saving them in standard \texttt{.csv} or \texttt{.json} formats. This architecture enables SCARF to efficiently assess and compare different RAG frameworks, assisting users in identifying the most suitable solution for their specific use case.
    \item \textbf{Ability to delve into metrics at different levels:} SCARF outputs per-question results, including text responses, expected answers, and metadata. An optional \texttt{EvaluatorGPT} module can measure correctness or consistency (using LLM-as-a-judge approach \cite{zheng2023judging}). Users may also integrate other specialized evaluators due to the modular nature of SCARF.
\end{itemize}

\subsection{Scenarios}
SCARF supports various testing scenarios to accommodate different user goals:

\subsubsection{ \textbf{Evaluating a Single RAG Framework}} 
Users may want to validate that a particular RAG framework correctly retrieves and generates answers from a given dataset. They may also experiment with different parameters (e.g. model temperature, retrieval thresholds, embedder, vectordb, LLM provider) to optimize performance for their own datasets. In this scenario, the system:
\begin{itemize}
    \item connect to the framework’s endpoint.
    \item uploads documents to the RAG knowledge base through the framework API.
    \item executes a set of queries (generic or file-specific) and saves responses.
    \item optionally runs an \texttt{evaluator} module (like \texttt{evaluator\_gpt.py}) to assess correctness, relevancy, or other NLP metrics.
\end{itemize}

\subsubsection{\textbf{Comparing Multiple RAG Frameworks on the Same Dataset}}
SCARF also supports comparisons among multiple frameworks (any RAG framework for which an adapter module is available or has been written by a user). SCARF runs identical test queries against each deployed framework in sequence, then it combines the results into a single \texttt{.csv} or \texttt{.json}. This simplifies questions like:
\begin{itemize}
     \item \textbf{Cross-framework analysis:} e.g. “Which RAG framework, with these specific settings, is the most accurate in domain X?' 
    \item \textbf{Performance benchmarks:} “Which approach yields the fastest response with the same hardware or number of documents?”
\end{itemize}

\section{Framework in Action}

This section provides a detailed look at how SCARF carries out its end-to-end evaluation processes in real-world scenarios. It highlights key points that facilitate the testing of multiple RAG platforms.

\subsection{SCARF workflow}
\begin{figure}[htbp]
\centering
\includegraphics[width=0.98\columnwidth]{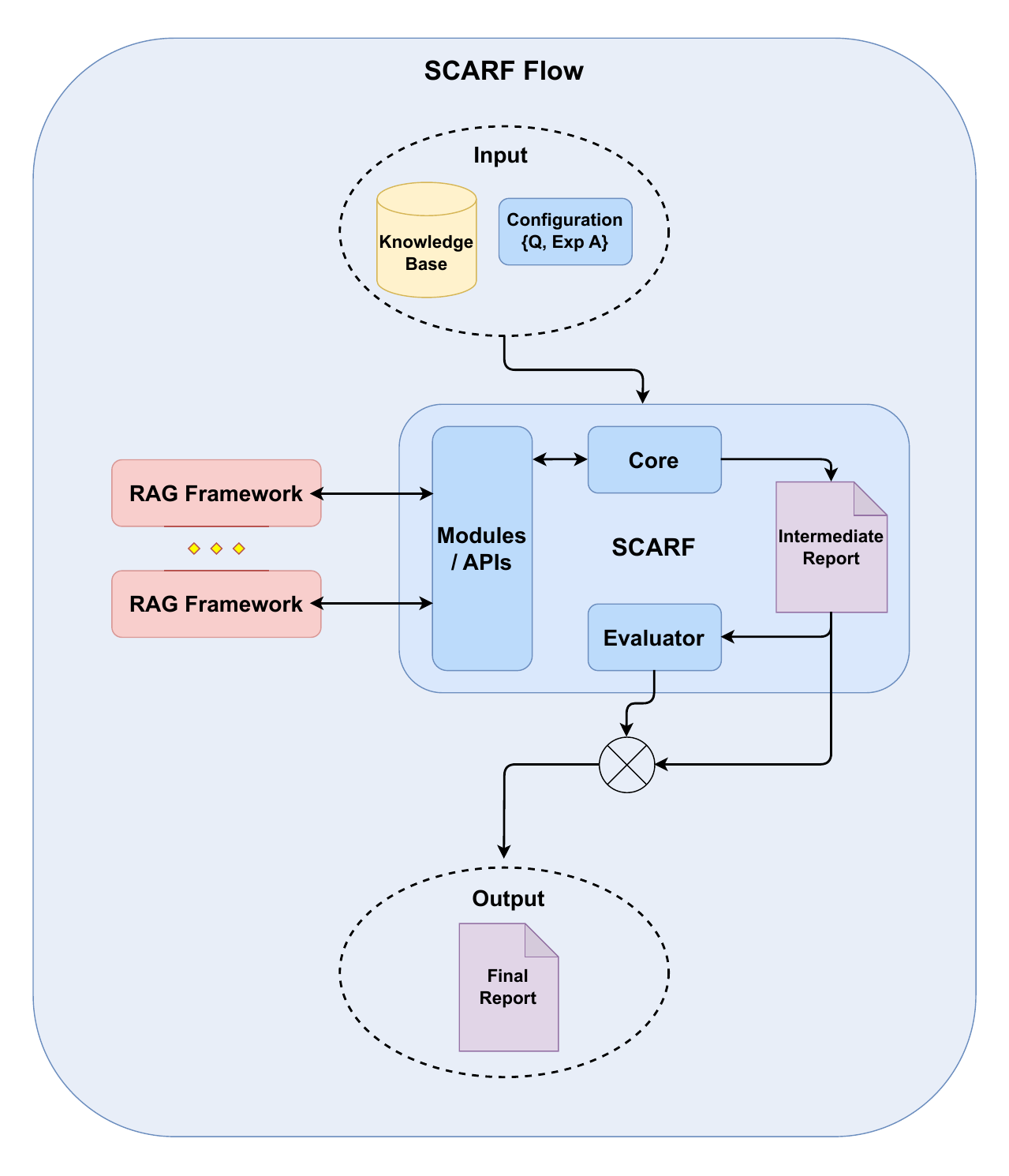}
    \caption{Flow showing how SCARF interact with data and RAG frameworks to produce the output.}
    \label{fig:scarf_flow}
\end{figure}

In Fig. \ref{fig:scarf_flow}, we show the workflow of our framework. The main entry point for SCARF evaluations is the script \texttt{test\_rag\_frameworks.py} which orchestrates all the procedures. Below, we summarize the high-level stages of SCARF’s operational flow:

\begin{enumerate}
    \item The system reads the input configuration to determine which files should be uploaded, identifies the specific queries that need to be executed, and selects the appropriate files from the knowledge base to ingest into the RAG platform. 
    \item Checking command-line flags (e.g. \texttt{--apikey}) to determine which framework(s) to target and any necessary credentials.
    \item Dynamically loading a corresponding API adapter (e.g. \texttt{CheshireCatAPI}, \texttt{AnythingLLMAPI}).
    \item Submitting queries to each RAG framework sequentially and collecting responses in memory via Modules/API.
    \item Optionally calling \texttt{EvaluatorGPT} to produce automated scores or annotations for each answer. Finally, SCARF saves both raw responses and computed evaluation metrics in standardized formats (e.g., csv).
    \item Output Export: \texttt{.csv}, \texttt{.json}. This step consolidates all data, ensures compatibility with common data processing tools (e.g., Pandas, Excel), and provides a complete snapshot of the experimental run.
\end{enumerate}

Algorithm~\ref{alg:scarf} presents a more detailed and comprehensive description of these steps in the form of pseudocode, providing further clarity on the procedural aspects involved.

\subsection{Writing an adapter Module for a Custom RAG Framework}
SCARF is designed to be \emph{highly extensible}, recognizing that practitioners may need to evaluate emerging or proprietary RAG solutions. Developers can integrate any system by creating a SCARF-compliant adapter module and placing it in the \texttt{modules/} directory. Each adapter must implement two key methods:

\begin{itemize}
    \item \texttt{upload\_document(file\_path: str) -> Dict[str, Any]}:
    Responsible for adding a local file to the framework’s knowledge base. In some RAG systems, this may involve splitting the file into smaller chunks before embedding and indexing. SCARF captures any returned metadata (e.g., document IDs, potential error messages).

    \item \texttt{send\_message(message: str) -> Dict[str, Any]}:
    Submits a text-based query to the RAG endpoint and captures both the raw text response and any auxiliary diagnostic data (e.g., top-ranked document identifiers, partial token sequences).
\end{itemize}

After placing this new module (e.g. \texttt{mynewrag\_api.py}) in the \texttt{modules/} folder, you can specify it in \texttt{test\_rag\_frameworks.py} or via command-line flags.
After integrating your custom module, you can interact with your deployed RAG platform and begin running tests, queries, and performance evaluations. Moreover, once you have refined your queries and expected responses, you have the chance to modify default metrics.

\mycomment{
\begin{algorithm}[ht]
\caption{SCARF Workflow}
\label{alg:scarf}
\begin{algorithmic}
\State \textbf{Input:} Configuration file (\texttt{config.json}), command-line arguments, and optional API keys.
\State \textbf{Output:} Evaluation results saved as CSV and JSON files.
\State \textbf{Begin:}
\State Load configuration from \texttt{config.json}.
\State Parse command-line flags for API keys, target framework(s), and test parameters.
\FORALL{selected RAG frameworks}
    \State Dynamically load the corresponding API adapter.
    \FORALL{warmup queries (not associated with a document)}
        \State response $\gets$ \texttt{send\_message(warmup\_query)}
        \State Append the warmup response and metadata to the results list.
    \ENDFOR
    \FORALL{each document in the dataset}
        \State \texttt{upload\_document(document\_path)}
        \FORALL{queries associated with the current document}
            \State response $\gets$ \texttt{send\_message(query)}
            \State Append the response and metadata to the results list.
        \ENDFOR
    \ENDFOR
\ENDFOR
\IF{evaluation mode is enabled}
    \State Call \texttt{EvaluatorGPT} to generate scores for each response.
    \State Merge evaluation scores with the raw responses.
\ENDIF
\State Save the aggregated results to \texttt{test\_results.csv} and \texttt{test\_results.json}.
\STATE \textbf{End.}
\end{algorithmic}
\end{algorithm}
}

\begin{algorithm}[ht]
\caption{SCARF Workflow}
\label{alg:scarf}
\begin{algorithmic}[1]
\STATE \textbf{Input:} Configuration file (\texttt{config.json}), command-line arguments, and optional API key.
\STATE \textbf{Output:} Evaluation results saved as CSV and JSON files.
\STATE \textbf{Begin:}
\STATE Load configuration from \texttt{config.json}.
\FORALL{selected RAG frameworks}
    \STATE Dynamically load the corresponding API adapter.
    \FORALL{warmup queries (not associated with a document)}
        \STATE response $\gets$ \texttt{send\_message(warmup\_query)}
        \STATE Append the warmup response and metadata to the results list.
    \ENDFOR
    \FORALL{document in the dataset}
        \STATE \texttt{upload\_document(document\_path)}
        \FORALL{queries associated with the current document}
            \STATE response $\gets$ \texttt{send\_message(query)}
            \STATE Save the response
        \ENDFOR
    \ENDFOR
\ENDFOR
\IF{evaluation mode is enabled}
    \STATE Call \texttt{EvaluatorGPT} for each response.
    \STATE Merge evaluation scores with the raw responses.
    \STATE Save the aggregated results to \texttt{test\_results.csv} or \texttt{test\_results.json}.
    \ENDIF
\STATE \textbf{End.}
\end{algorithmic}
\end{algorithm}

\section{Conclusion and Limitations}

In this technical report, we presented SCARF, a highly flexible and modular evaluation framework for RAG systems. Unlike many existing tools, which often focus on single components or assume local integration, SCARF operates at a “black-box” level. It can connect to any already-deployed RAG solution through a minimal adapter module, making it easy for researchers and practitioners to \textbf{assess multiple RAG frameworks} side by side on the same dataset, enabling direct comparisons of many performance indicators.

Our approach complements the capabilities of existing RAG evaluation frameworks, which may focus more narrowly on metrics for retrieval or generation. SCARF allows users to benchmark a range of real-world scenarios—from single-framework tuning to large-scale, multi-framework using a single, consistent interface. This modularity is particularly valuable in environments where organizations need to validate not only the quality of the model but also determine which RAG framework is the most suitable for their specific use case and data.

Despite its flexibility, SCARF has some limitations that could be addressed in future iterations. Currently, SCARF requires users to manually provide queries for evaluation, which can be time-consuming and may introduce biases in the assessment process. A promising direction for improvement is the integration of synthetic query generation, allowing SCARF to create diverse test cases and reducing human intervention.

Additionally, while SCARF supports a range of metrics, future versions could incorporate additional evaluation criteria, such as system response time, latency, stability under load, and scalability, to provide a more holistic performance assessment of different RAG frameworks.

Another area for improvement is the visualization and user experience. Currently, SCARF primarily focuses on providing numerical results and logs. The development of a graphical user interface (GUI) and a dedicated control panel would greatly enhance the user experience, making it easier to navigate through results, compare frameworks visually, and explore detailed insights across multiple experiments.

As the RAG landscape continues to evolve, we believe SCARF’s ability to integrate seamlessly with emerging tools will remain a key advantage. By addressing these limitations and expanding its capabilities, SCARF can become an even more comprehensive and user-friendly framework for evaluating RAG systems.

\bibliographystyle{IEEEtran}
\bibliography{bibliography}
\end{document}